\documentclass[10pt,twocolumn,letterpaper]{article}

\usepackage{iccv}
\usepackage{times}
\usepackage{epsfig}
\usepackage{graphicx}
\usepackage{amsmath}
\usepackage{amssymb}
\usepackage{bm}

\newcommand*{\rd}{\textcolor{red}}
\newcommand*{\gr}{\textcolor{green}}
\newcommand*{\bl}{\textcolor{blue}}

\usepackage[pagebackref=true,breaklinks=true,letterpaper=true,colorlinks,bookmarks=false]{hyperref}

\iccvfinalcopy 

\def\iccvPaperID{5235} 

\ificcvfinal\fi
\begin{document}

\title{Joint Learning of Saliency Detection and \\Weakly Supervised Semantic Segmentation}
\author{Yu Zeng, Yunzhi Zhuge, Huchuan Lu\thanks{Corresponding author.}, Lihe Zhang\\
Dalian University of Technology, China\\
{\tt\small \{zengyu, zgyz\}@mail.dlut.edu.cn, \{lhchuan, zhanglihe\}@dlut.edu.cn, }\\
}


\maketitle

\begin{abstract}
Existing weakly supervised semantic segmentation (WSSS) methods usually utilize the results of pre-trained saliency detection (SD) models without explicitly modelling the connections between the two tasks, which is not the most efficient configuration. Here we propose a unified multi-task learning framework to jointly solve WSSS and SD using a single network, \ie saliency and segmentation network (SSNet). SSNet consists of a segmentation network (SN) and a saliency aggregation module (SAM). For an input image, SN generates the segmentation result and, SAM predicts the saliency of each category and aggregating the segmentation masks of all categories into a saliency map. The proposed network is trained end-to-end with image-level category labels and class-agnostic pixel-level saliency labels. Experiments on PASCAL VOC 2012 segmentation dataset and four saliency benchmark datasets show the performance of our method compares favorably against state-of-the-art weakly supervised segmentation methods and fully supervised saliency detection methods. 
\end{abstract}

\section{Introduction}
Semantic image segmentation is an important and challenging task of computer vision, of which the goal is to predict a category label for every image pixel. Recently, convolutional neural networks (CNNs) have achieved remarkable success in semantic image segmentation~\cite{long2015fully, Ding_2019_CVPR, ding2018context, lin2017refinenet, chen2017deeplab, ding2019boundary}. Due to the expensive cost for annotating semantic segmentation labels to train CNNs, weakly supervised learning has attracted increasing interest, resulting in various weakly supervised semantic segmentation (WSSS) methods. 
Saliency detection (SD) aims at identifying the most distinct objects or regions in an image, which has helped many computer vision tasks such as scene classification~\cite{siagian2007rapid}, image retrieval~\cite{he2012mobile}, visual tracking~\cite{mahadevan2013biologically}, to name a few. With the success of deep CNNs, it has been made a lot of attempts to use deep CNNs or deep features for saliency detection~\cite{feng2019attentive,zeng2018learning,zeng2019multi,ZhugeZL19,zeng2018unsupervised,kong2018exemplar,fan2019shifting,fan2018salient}.

The two tasks both require to generate accurate pixel-wise masks. Hence, they have close connections. On the one hand, given the saliency map of an image, the computation load of a segmentation model can be reduced because of avoiding processing background. On the other hand, given the segmentation result of an image, the saliency map can be readily derived by selecting the salient category. 
Therefore, many existing WSSS~\cite{kolesnikov2016seed, wei2017object, wei2017stc, oh2017exploiting, huang2018weakly, wei2018revisiting, wang2018weakly, fan2018associating} methods have greatly benefited from SD. It is a widespread practice to exploit class activation maps (CAMs)~\cite{zhou2016learning} for locating the objects of each category and use SD methods for selecting background regions. 
For example, Wei~\etal~\cite{wei2018revisiting} use CAMs of a classification network with different dilated convolutional rates to find object regions, and use saliency maps of~\cite{xiao2017self} to find background regions for training a segmentation model. Wang~\etal~\cite{wang2018weakly} use saliency maps of ~\cite{jiang2013salient} to refine object regions produced by classification networks. 

However, those WSSS methods simply utilize the results of pre-trained saliency detection models, which is not the most efficient configuration. On the one hand, they use SD methods as a pre-processing step to generate annotations for training their segmentation models while ignoring the interactions between SD and WSSS, which blocks the WSSS models from fully exploiting the segmentation cues of the strong saliency annotations. On the other hand, heuristic rules are usually required for selecting background regions according to the results of SD models, thereby complicating the training process and leading to a not end-to-end manner. 
\begin{figure}[t]
\begin{center}
  \includegraphics[width=.9\linewidth]{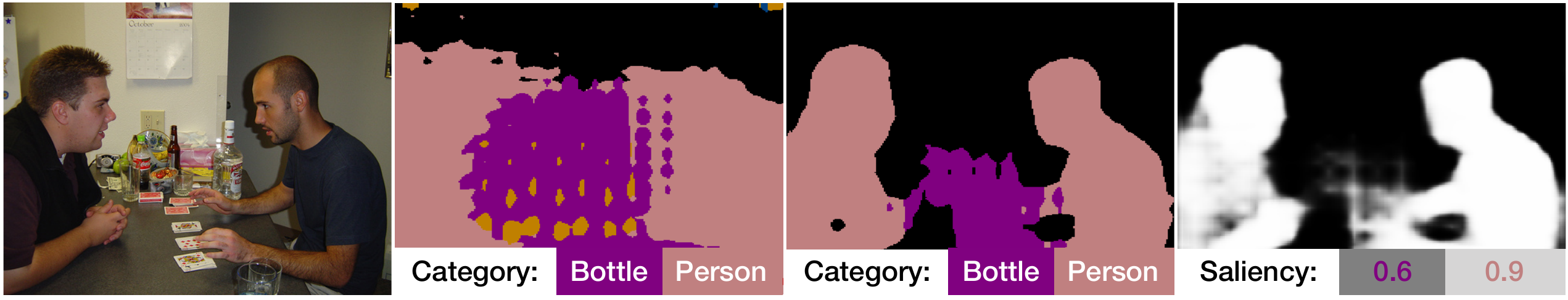}\\
  (a)~~~~~~~~~~~~~~(b)~~~~~~~~~~~~~~~~(c)~~~~~~~~~~~~~~~~(d)
\end{center}
  \caption{(a) input image. (b) segmentation results predicted by the model trained with only image-level labels. (c) segmentation results predicted by our method. (d) saliency map predicted by our method. }
\label{fig1}
\end{figure}

In this paper, we propose a unified, end-to-end training framework to solve both SD and WSSS tasks jointly. Unlike most existing WSSS methods that used pre-trained saliency detection models, we directly take advantage of pixel-level saliency labels. The core motive is to utilize semantic information of the image-level category labels and the segmentation cues of the category-agnostic saliency labels. The image-level category labels can make a CNN recognize the semantic categories, but they do not contain any spatial information, which is essential for segmentation. Although it has been suggested that CNNs trained with image-level labels are also informative of object locations, only a coarse spatial distribution can be inferred, as shown in the first row of Figure~\ref{fig1}. We solve this problem with the pixel-level saliency labels. Through explicitly modelling the connection between SD and WSSS, we derive the saliency maps from the segmentation results and minimize the loss between them and the saliency ground-truth. So that the CNN has to precisely cut the recognized objects so as to make the derived saliency maps match the ground-truth. 

Specifically, we propose a saliency and segmentation network (SSNet), which includes a segmentation network (SN) and a saliency aggregation module (SAM). For an input image, SN generates the segmentation results, as shown in The second column of Figure~\ref{fig1}. SAM predicts the saliency score of each category and then aggregates the segmentation masks of all categories into a saliency map according to their saliency scores, which bridges the gap between semantic segmentation and saliency detection. As shown in the third column of Figure~\ref{fig1}, given the segmentation map and saliency score of each category, saliency detection result can be generated by highlighting the masks of salient objects (\eg, the mask of persons in the third column) and suppressing the masks of the objects of low salience (\eg, the mask of bottles in the third column). When training, the loss is computed between the segmentation results and the image-level category labels as well as the saliency maps and the saliency ground-truth. 

Our approach has several advantages. First, compared with existing WSSS methods that exploit pre-trained SD models for pre-processing, our method explicitly models the relationships between saliency and segmentation, which can 
transfer the learned segmentation knowledge from class-agnostic image-specific saliency categories with pixel-level annotations to unseen semantic categories with only image-level annotations.
Second, as a low-level vision task, annotating pixel-level ground truth for saliency detection is less expensive than semantic segmentation. 
Therefore, compared with fully supervised segmentation methods, our method is trained with image-level category labels and saliency annotations, requiring less labeling cost. 
Third, compared with existing segmentation or saliency methods, our method can simultaneously predict the segmentation results and saliency results using a single model, with most parameters shared between the two tasks. 

In summary, our main contributions are three folds:
\begin{itemize}
\setlength{\itemsep}{0pt}
\setlength{\parsep}{0pt}
\setlength{\parskip}{0pt}
\item We propose a unified end-to-end framework for both SD and WSSS tasks, in which segmentation is split into two learning tasks respectively based on image-level category labels and pixel-level saliency annotations. 
\item We design a saliency aggregation module to explicitly bridge the two tasks, through which WSSS can directly benefit from saliency inference and vice versa. 
\item The experiments on the PASCAL VOC 2012 segmentation benchmark and four saliency benchmarks demonstrate the effectiveness of the proposed method. It achieves favorable performance against weakly supervised semantic segmentation methods and fully supervised saliency detection methods. We make our code and models available for further researches\footnote{https://github.com/zengxianyu/jsws}\footnote{http://ice.dlut.edu.cn/lu/}. 
\end{itemize}

\section{Related work}
\subsection{Saliency detection}
Earlier saliency detection methods used low-level features and heuristic priors~\cite{yang2013saliency, jiang2013submodular} to detect salient objects, which were not robust to complex scenes. Recently, deep learning based methods have achieved remarkable performance improvements. Incipient deep learning based methods usually used regions as computation units, such as superpixels, image patches, and region proposals. Wang~\etal~\cite{wang2015deep} trained two neural networks that estimate saliency of image patches and regional proposals respectively. Li and Yu~\cite{li2015visual} used CNNs to extract multi-scale features and predict the saliency of each superpixel. Inspired by the success of fully convolutional network (FCN)~\cite{long2015fully} on semantic segmentation, some methods have been proposed to exploit fully convolutional structure for pixel-wise saliency prediction. Liu and Han~\cite{liu2016dhsnet} proposed a deep hierarchical network to learn a coarse global saliency map and then progressively refine it. Wang~\etal~\cite{wang2016saliency} proposed a recurrent FCN incorporates saliency priors. Zhang~\etal~\cite{zhang2017learning} propose to make CNNs learn deep uncertain convolutional features (UCF) to encourage the robustness and accuracy of saliency detection. Zhang~\etal~\cite{zhang2018progressive} proposed an attention guided network which selectively integrates multi-level contextual information in a progressive manner. Chen~\etal~\cite{chen2018eccv} proposed reverse attention to guide residual feature learning in a top-down manner for saliency detection. All of the above saliency detection methods trained fully supervised models for a single task.  Although our method slightly increases labeling cost, it achieves state-of-the-art performance in both saliency detection and semantic segmentation. 

\subsection{Segmentation with weak supervision}
In recent years, a lot of weakly supervised semantic segmentation methods have been proposed to alleviate the cost of labeling. Various supervision has been exploited, such as the image-level labels, bounding boxes, scribbles, \etc. Among all kinds of weak supervision, the weakest one, \ie, image-level supervision, has attracted the most attention. In image-level weakly supervised segmentation, some methods exploited results of the pre-trained saliency detection models. A simple-to-complex method was presented in~\cite{wei2017stc}, in which an initial segmentation model is trained with simple images using saliency maps for supervision. Then the ability of the segmentation model is enhanced by progressively including samples of increasing complexity. Wei~\etal~\cite{wei2017object} iteratively used CAM~\cite{zhou2016learning} to discover object regions and used saliency detection results of~\cite{jiang2013salient} to find background regions to train the segmentation model.  Oh~\etal~\cite{oh2017exploiting} used an image classifier to find the high confidence points over the objects classes, \ie object seeds, and exploit a CNN-based saliency detection model to find the masks corresponding to some of the detected object seeds. Then these class-specific masks were used to train a segmentation model. Wei~\etal~\cite{wei2018revisiting} used a classification network with convolutional blocks of different dilated rates to find object regions and used saliency detection results of~\cite{xiao2017self} to find background regions to train a segmentation model. Wang~\etal~\cite{wang2018weakly} started from the object regions produced by classification networks. The object regions were expanded using the mined features and refined using saliency maps produced by~\cite{jiang2013salient}. Then the refined object regions were used as supervision to train a segmentation network. The above weakly supervised segmentation methods all exploited results of pre-trained saliency detection models, either using the existing models or separately training their saliency models and segmentation models. 
The proposed method has two main differences from these methods. First, these methods used pre-trained saliency detection models, while we directly exploit strong saliency annotations and work in an end-to-end manner. Second, in these methods, saliency detection was used as a pre-processing step to generate training data for segmentation. In contrast, we simultaneously solve saliency detection and semantic segmentation using a single model, of which most parameters are shared between the two tasks. 
   
\subsection{Multi-task learning}
Multi-task learning has been used in a wide range of computer vision problems. Teichman~\etal~\cite{teichmann2018multinet} proposed an approach to joint classification, detection, and segmentation using a unified architecture where the encoder is shared among the three tasks. Kokkinos~\cite{kokkinos2017ubernet} proposed an UberNet that jointly handles low-, mid-, high-level tasks including boundary detection, normal estimation, saliency estimation, semantic segmentation, human part segmentation, semantic boundary detection, region proposal generation, and object detection. Eigen and Fergus~\cite{eigen2015predicting} used a multiscale CNN to address three different computer vision tasks: depth prediction, surface normal estimation, and semantic labeling. Xu~\etal~\cite{xu2018pad} proposed a PAD-Net that first solves several auxiliary tasks ranging from low level to high level, and then used the predictions as multi-modal input for the final task. The models above all worked in full supervision setting. In contrast, we jointly learn to solve a task in weak supervision setting and another task in full supervision setting.

\section{The proposed approach}
\begin{figure*}[t]
\begin{center}
\includegraphics[width=.9\linewidth]{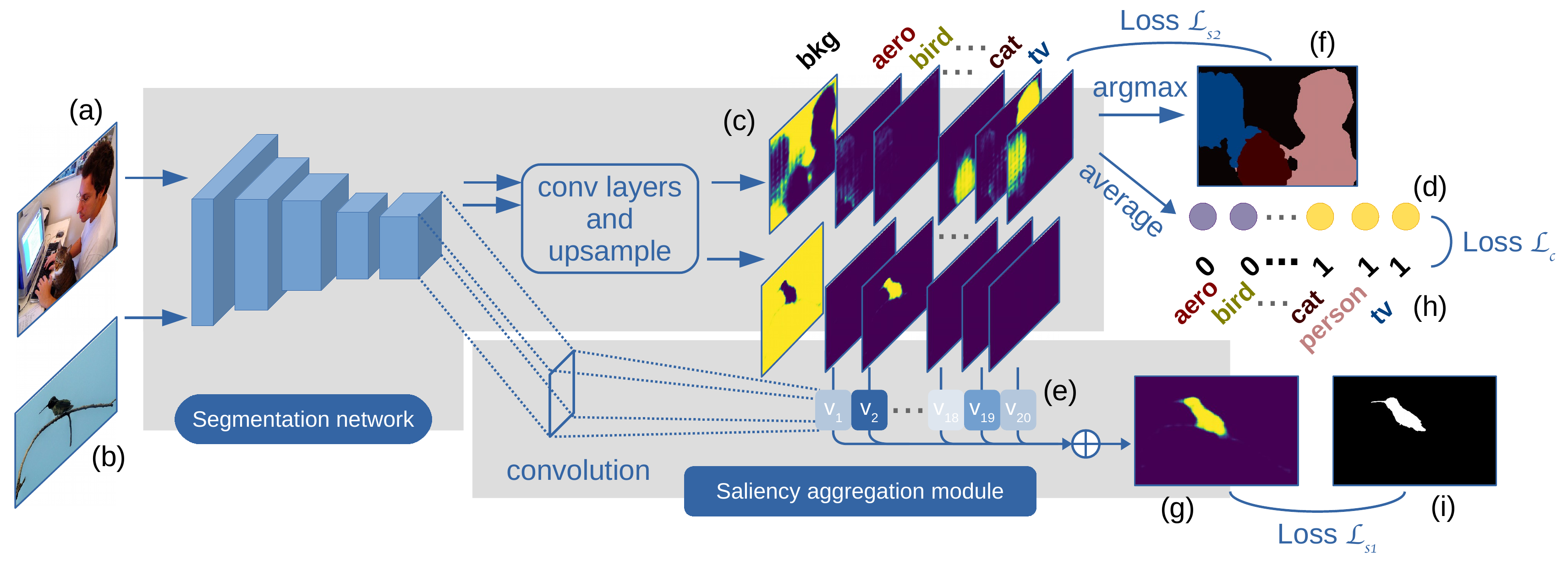}
\end{center}
  \caption{An overview of the proposed method. Our model is trained with (a) \textit{images annotated with category labels} and (b) \textit{images with saliency ground-truth}. For an input image, the segmentation network generates a (c) \textit{segmentation results}, of which the average over spatial locations indicates (d) \textit{the probability of each category}. The saliency aggregation module predicts (e) \textit{saliency score of each category} to aggregate segmentation masks of all categories into a (g) \textit{saliency map}. In the first training stage, the network is trained with the (h) \textit{category labels} and (i) \textit{saliency ground truth}. In the second training stage, the network is trained with (f) \textit{predicted segmentation results} by the model trained in the first stage and the saliency ground truth.}
\label{fig3}
\end{figure*}
In this section, we detail the joint learning framework for simultaneous saliency detection and semantic segmentation. We first give an overview of the proposed saliency and segmentation network (SSNet). Then we describe the details of the segmentation network (SN) and the saliency aggregation module (SAM) in Section~\ref{sec_sn} and~\ref{sec_pum_sam}. Finally, we present the joint learning strategy in Section~\ref{sec_train}. Figure~\ref{fig3} illustrates the overall architecture of the proposed method.

\subsection{Network overview}
We design two variants of SSNet, \ie SSNet-1, and SSNet-2, for two training stages, respectively. In the first training stage, the SSNet-1 is trained with pixel-level saliency annotations and image-level semantic category labels. In the second stage, the SSNet-2 is trained with saliency annotations and image-level semantic category labels as well as semantic segmentation results predicted by SSNet-1. Both the SSNet-1 and SSNet-2 consist of a segmentation network (SN) and a saliency aggregation module (SAM). Given an input image, SN predicts a segmentation result. SAM predicts a saliency score for each semantic class and aggregates the segmentation map into a single channel saliency map according to the saliency score of each class. Both the SSNet-1 and SSNet-2 are trained end-to-end.  
\subsection{Segmentation networks}\label{sec_sn}
The segmentation network consists of a feature extractor to extract features from the input image and several convolution layers to predict segmentation results given the features. Feature extractors of our networks are designed based
on state-of-the-art CNN architectures for image recognition, \eg, VGG~\cite{karenvery} and DenseNet~\cite{huang2017densely}, which typically contain five convolutional blocks for feature extraction and a fully connected classifier. We remove the fully connected classifier and use the convolutional blocks as our feature extractor. To obtain larger feature maps, we remove the downsampling operator from the last two convolution blocks and use dilated convolution to retain the original receptive field. The feature extractor generates feature maps of $1/8$ the input image size. We resize the input images to $256\times 256$, so the resulted feature maps are $32\times 32$ in spatial scale.

In the first training stage, the only available semantic supervision cue is the image-level labels. Trained with image-level labels, a coarse spatial distribution of each class can be inferred but it is difficult to train a sophisticated model. Therefore, we use a relatively simple structure in SSNet-1 for generating segmentation results, \ie a $1\times 1$ convolution layer. The predicted $C$-channel segmentation map and one-channel background map are of $1/8$ the input image size, in which $C$ is the number of semantic classes. Each element of the segmentation map and background map is a value in $[0, 1]$. The values of all classes sum to 1 for each pixel. Then the segmentation results are upsampled by a deconvolution layer to the input image size. In the second training stage, the segmentation results of SSNet-1 can be used for training, which is a stronger supervision cue. Therefore, we can use a more complex segmentation network to generate finer segmentation results. Inspired by Deeplab~\cite{chen2018deeplab}, we use four $3\times 3$ convolution layers with dilation rate $6, 12, 18, 24$ in SSNet-2 and take the summation of their outputs as the segmentation results. Similar to SSNet-1, these segmentation results are of $1/8$ the input image size, and are upsampled by a deconvolution layer to the input size. 


\subsection{Saliency aggregation}\label{sec_pum_sam}
We design a saliency aggregation module (SAM) as a bridge between the two tasks so that the segmentation network can make use of the class-agnostic pixel-level saliency labels and generate more accurate segmentation results. This module takes the $32\times 32$ outputs $F$ of the feature extractor, and generates a $C$-dimensional vector $\bm{v}$ with a $32\times 32$ convolution layer and a sigmoid function, of which each element $v_i$ is the saliency score of the $i$-th category. Then the saliency map $S$ is given by a weighted sum of the segmentation masks of all classes:
\begin{equation}
S = \sum_{i=1}^C v_i \cdot H_i, 
\end{equation}
where $H_i$ denotes the $i$-th channel of the segmentation results encoding the spatial distribution of the $i$-th category, which are the output of the segmentation network. 

\subsection{Jointly learning of saliency and segmentation}~\label{sec_train}
We use two training sets to train the proposed SSNet: the saliency dataset with pixel-level saliency annotations, and the classification dataset with image-level semantic category labels. Let $\mathcal{D}_s = \{ (X^n, Y^n)\}_{n=1}^{N_s}$ denote the saliency dataset, in which $X^n$ is the image, and $Y^n$ is the ground truth. Each element of $Y^n$ is either 1 or 0, representing the corresponding pixel belongs to salient objects or background, respectively. The classification dataset is denoted as $\mathcal{D}_c = \{ (X^n, \bm{t}^n)\}_{n=1}^{N_c}$, in which $X^n$ is the image, and $\bm{t}^n$ is the one-hot encoding of the categories of the image. 

For an input image, the segmentation network generates its segmentation result, from which the probability of each category can be derived by averaging the segmentation results over spatial locations. We compute the loss between these values and the ground-truth category labels and backward propagate it to make the segmentation results semantically correct, \ie, the semantic categories appearing in the input image are correctly recognized. This loss, denoted as $\mathcal{L}_c$, is defined as follow, 
\begin{equation}
\begin{split}
\mathcal{L}_c &= -\frac{1}{N_c}\sum_{n=1}^{N_c} \left[ \sum_{i=1}^C t_i^n \log \hat{t}_i^n + (1-t_i^n) \log (1-\hat{t}_i^n) \right],\\
\end{split}
\end{equation}
in which $t_i^n$ is the $i$-th element of $\bm{t}^n$. $t_i^n=1$ represents the image $X^n$ contains objects of the $i$-th category, and $t_i^n=0$ otherwise. $\bm{\hat{t}}^n$ is the average over spatial positions of the segmentation maps $H^n$ of image $X^n$, of which each element $\hat{t}^n_i \in [0, 1]$ represents the predicted probability of the $i$-th class objects presenting in the image. 

The image-level category labels can make the segmentation network recognize the semantic categories, but they do not contain any spatial information, which is essential for segmentation. We solve this problem with the pixel-level saliency labels. As stated in Section~\ref{sec_pum_sam}, the SAM generates the saliency score of each category and aggregates the segmentation result into a saliency map. We minimize a loss $\mathcal{L}_{s1}$ between the derived saliency maps and the ground-truth so that the segmentation network has to precisely cut the recognized objects to make the derived saliency maps match the ground-truth. The loss $\mathcal{L}_{s1}$ between the saliency maps and the saliency ground-truth is defined as follow,
\begin{equation}
\begin{split}
\mathcal{L}_{s1} &= -\frac{1}{N_s}\sum_{n=1}^{N_s}\left[\sum_m y_m^n \log s_m^n + (1-y_m^n) \log (1-s_m^n)  \right], \\
\end{split}
\end{equation}
where $y_m^n \in \{0, 1\}$ is the value of the $m$-th pixel of the saliency ground truth $Y^n$. $s_m^n \in [0, 1]$ is the value of the $m$-th pixel in the saliency map of the image $X^n$, encoding the predicted probability of the $m$-th pixel being salient.

In the first training stage, we train SSNet-1 with the loss $\mathcal{L}_c+\mathcal{L}_{s1}$. After having trained SSNet-1, we run it on the classification dataset $\mathcal{D}_c$ and obtain the $C+1$-channel segmentation results, of which the first $C$ channels correspond to the $C$ semantic categories and the last channel corresponds to the background. Then the first $C$ channels of the segmentation result are cross-channel multiplied with the one-hot class label $\bm{t}^n$ to suppress wrong predictions and refined with CRF~\cite{krahenbuhl2011efficient} to enhance spatial smoothness. Finally, we obtain some pseudo labels by assigning each pixel $m$ of each training image $X^n \in \mathcal{D}_c$ a class label including the background label corresponding to the maximum value in the refined segmentation result. We define a loss $\mathcal{L}_{s2}$ between the segmentation results and the pseudo labels as follow, 
\begin{equation}
\begin{split}
\mathcal{L}_{s2} &= -\frac{1}{N_c}\sum_{n=1}^{N_c} \left[ \sum_{i=1}^{C+1} \sum_{m\in \mathcal{C}_i} \log h^n_{im}\right],\\
\end{split}
\end{equation}
in which $h^n_{im} \in [0, 1], i=1, ..., C$ is the value of $H^n$ at pixel $m$ and channel $i$, representing the probability of pixel $m$ belonging to the $i$-th class. SSNet-2 is trained with the loss $\mathcal{L}_{s1}+\mathcal{L}_{s2}$.

\section{Experiments}
\subsection{Dataset and settings}
\noindent
\textbf{Segmentation} For semantic segmentation task, we evaluate the proposed
method on the PASCAL VOC 2012 segmentation benchmark~\cite{everingham2015pascal}. This dataset has 20 object categories and one background category. It is split into a training set of 1,464 images, a validation set of 1,449 images and a test set of 1,456 images. Following the common practice~\cite{chen2014semantic, hariharan2011semantic, wei2017object}, we increase the number of training images to 10,582 by augmentation. We only use image-level labels for training. The performance of our method and other state-of-the-art methods are evaluated on the validation set and test set. The performance for semantic segmentation is evaluated in terms of inter-section-over-union averaged over 21 classes (mIOU) according to the PASCAL VOC evaluation criterion. We obtain the mIOU on the test set by submitting our results to the PASCAL VOC evaluation server. 

\noindent
\textbf{Saliency} For saliency detection task, we use the DUT-S training set~\cite{wang2017learning} for training, which has 10,553 images with pixel-level saliency annotations. The proposed method and other state-of-the-art methods are evaluated on four benchmark datasets: ECSSD~\cite{yan2013hierarchical}, PASCAL-S~\cite{li2014secrets}, HKU-IS~\cite{li2015visual}, SOD~\cite{martin2001database}. 
ECSSD contains 1000 natural images with multiple objects of different sizes. PASCAL-S stems from the validation set of PASCAL VOC 2010 segmentation dataset and contains 850 natural images. HKU-IS has 4447 images chosen to include multiple disconnected objects or objects touching the image boundary. SOD has 300 challenging images, of which many images contain multiple objects either with low contrast or touching the image boundary. The performance for saliency detection is evaluated in terms of maximum F-measure and mean absolute error (MAE). 

\noindent
\textbf{Training/Testing Settings} We adopt DenseNet-169~\cite{huang2017densely} pre-trained on ImageNet~\cite{deng2009imagenet} as the feature extractor of our segmentation network due to its ability to achieve comparable performance with a smaller number of parameters than other architectures. Our network is implemented based on Pytorch framework and trained on two NVIDIA
GeForce GTX 1080 Ti GPU. We use Adam optimizer~\cite{kingma2014adam} to train our network. We randomly crop a patch of 9/10 of the original image size and rescaled to $256\times 256$ when training. The batch size is set to 16. We train both the SSNet-1 and SSNet-2 for 10,000 iterations with initial learning rate 1e-4, and decrease the learning rate by 0.5 every 1000 iterations. When testing, the input image is resized to $256\times 256$. Then, the predicted segmentation results and saliency maps are resized to the input size by nearest interpolation. We do not use any post-processing on the segmentation results. We apply CRF~\cite{krahenbuhl2011efficient} to refine the saliency maps. 

\subsection{Comparison with saliency methods}
\begin{figure}[t!]
\begin{center}
  \includegraphics[width=.9\linewidth]{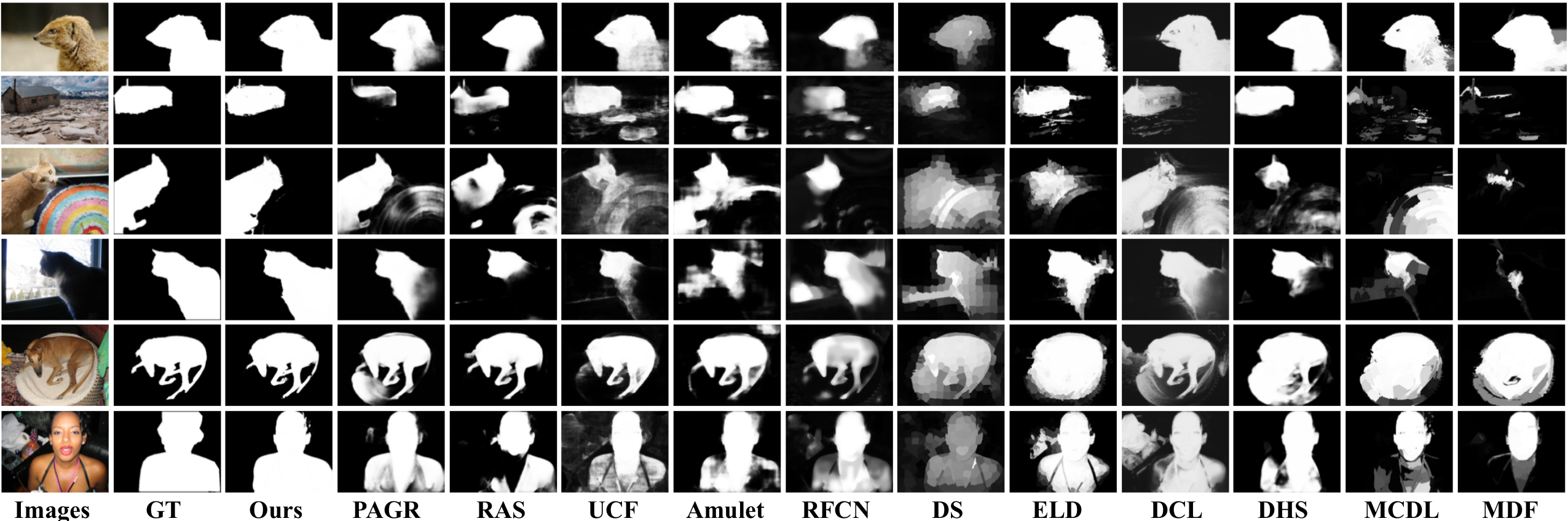}
\end{center}
  \caption{Visual comparison of the proposed method with state-of-the-art fully supervised saliency detection methods. }
\label{vis_sal}
\end{figure}

We compare our method with the following state-of-the-art deep learning based fully supervised saliency detection methods: PAGR (CVPR'18)~\cite{zhang2018progressive}, RAS (ECCV'18)~\cite{chen2018eccv}, UCF (ICCV'17)~\cite{zhang2017learning}, Amulet (ICCV'17)~\cite{zhang2017amulet}, RFCN (ECCV'16)~\cite{wang2016saliency}, DS (TIP'16)~\cite{li2016deepsaliency}, ELD (CVPR'16)~\cite{lee2016deep}, DCL (CVPR'16)~\cite{li2016deep}, DHS (CVPR'16)~\cite{liu2016dhsnet}, MCDL (CVPR'15)~\cite{zhao2015saliency}, MDF (CVPR'15)~\cite{li2015visual}. Figure~\ref{vis_sal} shows a visual comparison of our method against state-of-the-art fully supervised saliency detection methods. The comparison in terms of MAE and maximum F-measure is shown in Table~\ref{comp_mae} and Table~\ref{comp_fm} respectively. As shown in Table~\ref{comp_mae}, the proposed method achieves the smallest MAE among across all datasets. Maximum F-measure in Table~\ref{comp_fm} also shows that our method achieves the second largest F-measure in one dataset, and achieves the third largest F-measure in the other three datasets. Together the two metrics, it can be seen that our method achieves state-of-the-art performance in saliency detection task. 
\begin{table*}[t!]
\caption{\small Comparison of fully supervised saliency detection methods in terms of MAE (the smaller the better). The best three results are in \rd{red}, \gr{green} and \bl{blue}, respectively. }
\vspace{-10pt}
\label{comp_mae}
\small
\begin{center}
{\setlength{\tabcolsep}{.4em}
\begin{tabular}{c|cccccccccccc}
\hline
 Methods/Datasets &RAS\tiny '18&PAGR\tiny '18 &UCF \tiny '17& Amule\tiny '17& RFCN\tiny '16 & DS\tiny '16 & ELD\tiny '16 & DCL \tiny '16 & DHS\tiny '16 & MCDL \tiny '15 & MDF \tiny '15 &Ours\\
\hline
ECSSD &\gr{0.056} &\bl{0.061}   & 0.078& 0.059& 0.107& 0.122& 0.079& 0.088& 0.059& 0.101& 0.105     &\rd{0.045}\\
PASCAL-S &\bl{0.104} &\gr{0.093}  & 0.126& 0.098& 0.118& 0.176& 0.123& 0.125& 0.094& 0.145& 0.146    &\rd{0.067}\\
HKU-IS &\gr{0.045} &\bl{0.048}   & 0.074& 0.052& 0.079& 0.080& 0.074& 0.072& 0.053& 0.092& 0.129    &\rd{0.040}\\
DUTS-test &\bl{0.060} &\gr{0.056}  & 0.117& 0.085& 0.091& 0.090& 0.093& 0.088& 0.067& 0.106& 0.094  &\rd{0.052}\\
\hline
\end{tabular}}
\end{center}
\vspace{-10pt}
\end{table*}
\begin{table*}[t!]
\caption{\small Comparison of fully supervised saliency detection methods in terms of maximum F-measure (the larger the better). The best three results are in \rd{red}, \gr{green} and \bl{blue}, respectively.}
\vspace{-10pt}
\label{comp_fm}
\small
\begin{center}
{\setlength{\tabcolsep}{.4em}
\begin{tabular}{l|llllllllllll}
\hline
 Methods/Datasets &RAS\tiny '18&PAGR\tiny '18 &UCF \tiny '17& Amule\tiny '17& RFCN\tiny '16 & DS\tiny '16 & ELD\tiny '16 & DCL \tiny '16 & DHS\tiny '16 & MCDL \tiny '15 & MDF \tiny '15 &Ours\\
\hline
ECSSD &\gr{0.921}& \rd{0.927}& 0.911& 0.915& 0.890& 0.882& 0.867& 0.890& 0.907& 0.837& 0.832    &\bl{0.919}\\
PASCAL-S &\bl{0.837}& \rd{0.856}& 0.828& 0.837& 0.837& 0.765& 0.773& 0.805& 0.829& 0.743& 0.768   &\gr{0.851}\\
HKU-IS &\gr{0.913}&\rd{0.918}& 0.886& 0.895& 0.892& 0.865& 0.839& 0.885& 0.890& 0.808& 0.861    &\bl{0.907}\\
DUTS-test &\bl{0.831}&\rd{0.855}& 0.771& 0.778& 0.784& 0.777& 0.738& 0.782& 0.807& 0.672& 0.730  &\gr{0.832}\\
\hline
\end{tabular}}
\end{center}
\end{table*}
\vspace{0pt}

\subsection{Comparison with segmentation methods}
In this section, we compare our method with previous state-of-the-art weakly supervised semantic segmentation methods, \ie MIL (CVPR'15)~\cite{pinheiro2015image}, WSSL~\cite{papandreou2015weakly}, RAWK~\cite{vernaza2017learning}, BFBP (ECCV'16)~\cite{saleh2016built}, SEC (ECCV'16)~\cite{kolesnikov2016seed}, AE (CVPR'17)~\cite{wei2017object}, STC (PAMI'17)~\cite{wei2017stc}, CBTS (CVPR'17)~\cite{roy2017combining}, ESOS (CVPR'17)~\cite{oh2017exploiting}, MCOF (CVPR'18)~\cite{wang2018weakly}, MDC (CVPR'18)~\cite{wei2018revisiting}. WSSL uses bounding boxes as supervision, RAWK uses scribbles as supervision, and other methods use image-level categories as supervision. Among the methods using image-level supervision, ESOS exploits the saliency detection results of a deep CNN trained with bounding box annotations. AE, STC, MCOF, MDC use the results of fully supervised saliency detection models and thus implicitly use pixel level saliency annotations. As some previous methods used VGG16 as its backbone network, we also report the performance of our method using VGG16. It can be seen from Table~\ref{comp_val} and Table~\ref{comp_test} that our method compares favourably against all the above methods, including methods using stronger supervision such as bounding boxes (WSSL) and scribbles (RAWK). Our method also outperforms the methods \ie, ESOS, AE, STC, MCOF, MDC, that implicitly use saliency annotations by using pre-trained saliency detection models. Compared with these methods, our method simultaneously solves semantic segmentation and saliency detection and can be trained in an end-to-end manner, which is more efficient and easier to train. 
\begin{table*}[t!]
\caption{\small Comparison of WSSS methods on PASCAL VOC 2012 validation set. $^*$ marks the methods implicitly use saliency annotations by using pre-trained saliency detection models. $^\dag$ and $^\ddag$ mark the methods use box supervisions and scribble supervisions, respectively. Ours: our method with Densenet169-based feature extractor. Ours-VGG: our method with VGG16-based feature extractor. MCOF-Res: MCOF with ResNet101-based feature extractor. MCOF-VGG: MCOF with VGG16-based feature extractor. The best three results are in \rd{red}, \gr{green} and \bl{blue}.}
\vspace{-10pt}
\label{comp_val}
\small
\begin{center}
\begin{minipage}{\textwidth}
{\setlength{\tabcolsep}{.2em}
\begin{tabular}{c|ccccccccccccccccccccc|c}
\hline
Method & bkg & areo & bike & bird & boat & bottle & bus & car & cat & chair & cow & table & dog & horse & mbk & person & plant & sheep & sofa & train & tv & mean\\
\hline
MIL \tiny ’15 & 74.7 & 38.8 & 19.8 & 27.5 & 21.7 & 32.8 & 40.0 & 50.1 & 47.1 & 7.2 & 44.8 & 15.8 & 49.4 & 47.3 & 36.6 & 36.4 & 24.3 & 44.5 & 21.0 & 31.5 & 41.3 & 35.8\\
WSSL$^\dag$ \tiny '15&-&-&-&-&-&-&-&-&-&-&-&-&-&-&-&-&-&-&-&-&-&60.6\\
BFBP \tiny '16&79.2&60.1&20.4&50.7&41.2&46.3&62.6&49.2&62.3&13.3&49.7&38.1&58.4&49.0&57.0&48.2&27.8&55.1&29.6&54.6&26.6&46.6\\
SEC\tiny '16&82.2&61.7&26.0&60.4&25.6&45.6&70.9&63.2&72.2&20.9&52.9&30.6&62.8&56.8&63.5&57.1&32.2&60.6&32.3&44.8&42.3&50.7\\
RAWK$^\ddag$ \tiny '17&-&-&-&-&-&-&-&-&-&-&-&-&-&-&-&-&-&-&-&-&-&61.4\\
STC$^*$\tiny '17&84.5&68.0&19.5&60.5&42.5&44.8&68.4&64.0&64.8&14.5&52.0&22.8&58.0&55.3&57.8&60.5&40.6&56.7&23.0&57.1&31.2&49.8\\
AE$^*$\tiny '17&-&-&-&-&-&-&-&-&-&-&-&-&-&-&-&-&-&-&-&-&-&55.0\\
CBTS\tiny '17&85.8&65.2&29.4&63.8&31.2&37.2&69.6&64.3&76.2&21.4&56.3&29.8&68.2&60.6&66.2&55.8&30.8&66.1&34.9&48.8&47.1&52.8\\
ESOS$^*$\tiny '17&-&-&-&-&-&-&-&-&-&-&-&-&-&-&-&-&-&-&-&-&-&55.7\\
MCOF-Res$^*$\tiny '18&87.0&78.4&29.4&68.0&44.0&67.3&80.3&74.1&82.2&21.1&70.7&28.2&73.2&71.5&67.2&53.0&47.7&74.5&32.4&71.0&45.8&\bl{60.3}\\
MCOF-VGG$^*$\tiny'18&85.8&74.1&23.6&66.4&36.6&62.0&75.5&68.5&78.2&18.8&64.6&29.6&72.5&61.6&63.1&55.5&37.7&65.8&32.4&68.4&39.9&56.2\\
MDC$^*$\tiny '18&89.5&85.6&34.6&75.8&61.9&65.8&67.1&73.3&80.2&15.1&69.9&8.1&75.0&68.4&70.9&71.5&32.6&74.9&24.8&73.2&50.8&\gr{60.4}\\
Ours-VGG&89.1&71.5&31.0&74.2&58.6&63.6&78.1&69.2&74.4&10.7&63.6&9.8&66.4&64.4&66.6&64.8&27.5&69.2&24.3&71.0&50.9&57.1\footnote{http://host.robots.ox.ac.uk:8080/anonymous/F5E3DJ.html}\\
Ours&90.0&77.4&37.5&80.7&61.6&67.9&81.8&69.0&83.7&13.6&79.4&23.3&78.0&75.3&71.4&68.1&35.2&78.2&32.5&75.5&48.0&\rd{63.3}\footnote{http://host.robots.ox.ac.uk:8080/anonymous/AOZU76.html}\\
\hline
\end{tabular}}
\end{minipage}
\vspace{-10pt}
\end{center}
\end{table*}
\begin{table*}[t!]
\caption{\small Comparison of WSSS methods on PASCAL VOC 2012 test set. $^*$ marks the methods implicitly use saliency annotations by using pre-trained saliency detection models. $^\dag$ and $^\ddag$ mark the methods use box supervisions and scribble supervisions, respectively. Ours: our method with Densenet169-based feature extractor. Ours-VGG: our method with VGG16-based feature extractor. MCOF-Res: MCOF with ResNet101-based feature extractor. MCOF-VGG: MCOF with VGG16-based feature extractor. The best three are in \rd{red}, \gr{green}, \bl{blue}.}
\label{comp_test}
\small
\begin{center}
\begin{minipage}{\textwidth}
{\setlength{\tabcolsep}{.2em}
\begin{tabular}{c|ccccccccccccccccccccc|c}
\hline
Method & bkg & areo & bike & bird & boat & bottle & bus & car & cat & chair & cow & table & dog & horse & mbk & person & plant & sheep & sofa & train & tv & mean\\
\hline
MIL\tiny '15&74.7&38.8&19.8&27.5&21.7&32.8&40.0&50.1&47.1&7.2&44.8&15.8&49.4&47.3&36.6&36.4&24.3&44.5&21.0&31.5&41.3&35.8\\
WSSL$^\dag$ \tiny '15&-&-&-&-&-&-&-&-&-&-&-&-&-&-&-&-&-&-&-&-&-&62.2\\
BFBP\tiny '16&80.3&57.5&24.1&66.9&31.7&43.0&67.5&48.6&56.7&12.6&50.9&42.6&59.4&52.9&65.0&44.8&41.3&51.1&33.7&44.4&33.2&48.0\\
SEC\tiny '16&83.5&56.4&28.5&64.1&23.6&46.5&70.6&58.5&71.3&23.2&54.0&28.0&68.1&62.1&70.0&55.0&38.4&58.0&39.9&38.4&48.3&51.7\\
STC$^*$\tiny ’16&85.2&62.7&21.1&58.0&31.4&55.0&68.8&63.9&63.7&14.2&57.6&28.3&63.0&59.8&67.6&61.7&42.9&61.0&23.2&52.4&33.1&51.2\\
AE$^*$\tiny '17&-&-&-&-&-&-&-&-&-&-&-&-&-&-&-&-&-&-&-&-&-&55.7\\
CBTS\tiny '17&85.7&58.8&30.5&67.6&24.7&44.7&74.8&61.8&73.7&22.9&57.4&27.5&71.3&64.8&72.4&57.3&37.0&60.4&42.8&42.2&50.6&53.7\\
ESOS$^*$\tiny '17&-&-&-&-&-&-&-&-&-&-&-&-&-&-&-&-&-&-&-&-&-&56.7\\
MCOF-Res$^*$\tiny '18&88.2&80.8&31.4&70.9&34.9&65.7&83.5&75.1&79.0&22.0&70.3&31.7&77.7&72.9&77.1&56.9&41.8&74.9&36.6&71.2&42.6&\gr{61.2}\\
MCOF-VGG$^*$\tiny'18&86.8&73.4&26.6&60.6&31.8&56.3&76.0&68.9&79.4&18.8&62.0&36.9&74.5&66.9&74.9&58.1&44.6&68.3&36.2&64.2&44.0&57.6\\
MDC$^*$\tiny '18&89.8&78.4&36.2&82.1&52.4&61.7&64.2&73.5&78.4&14.7&70.3&11.9&75.3&74.2&81.0&72.6&38.8&76.7&24.6&70.7&50.3&\bl{60.8}\\
Ours-VGG&89.2&75.4&31.0&72.3&45.0&56.6&79.3&73.2&73.9&14.1&64.4&19.7&69.5&71.1&76.7&64.7&41.8&70.9&27.5&68.2&46.6&58.6\footnote{http://host.robots.ox.ac.uk:8080/anonymous/YTXEXK.html}\\
Ours &90.4&85.4&37.9&77.2&48.2&64.5&83.9&74.8&83.4&15.9&72.4&34.3&80.0&77.3&78.5&69.0&41.9&76.3&38.3&72.3&48.2&\rd{64.3}\footnote{http://host.robots.ox.ac.uk:8080/anonymous/PHYZSJ.html}\\
\hline
\end{tabular}}
\end{minipage}
\vspace{-10pt}
\end{center}
\end{table*}
\begin{figure}[t!]
\begin{center}
  \includegraphics[width=.9\linewidth]{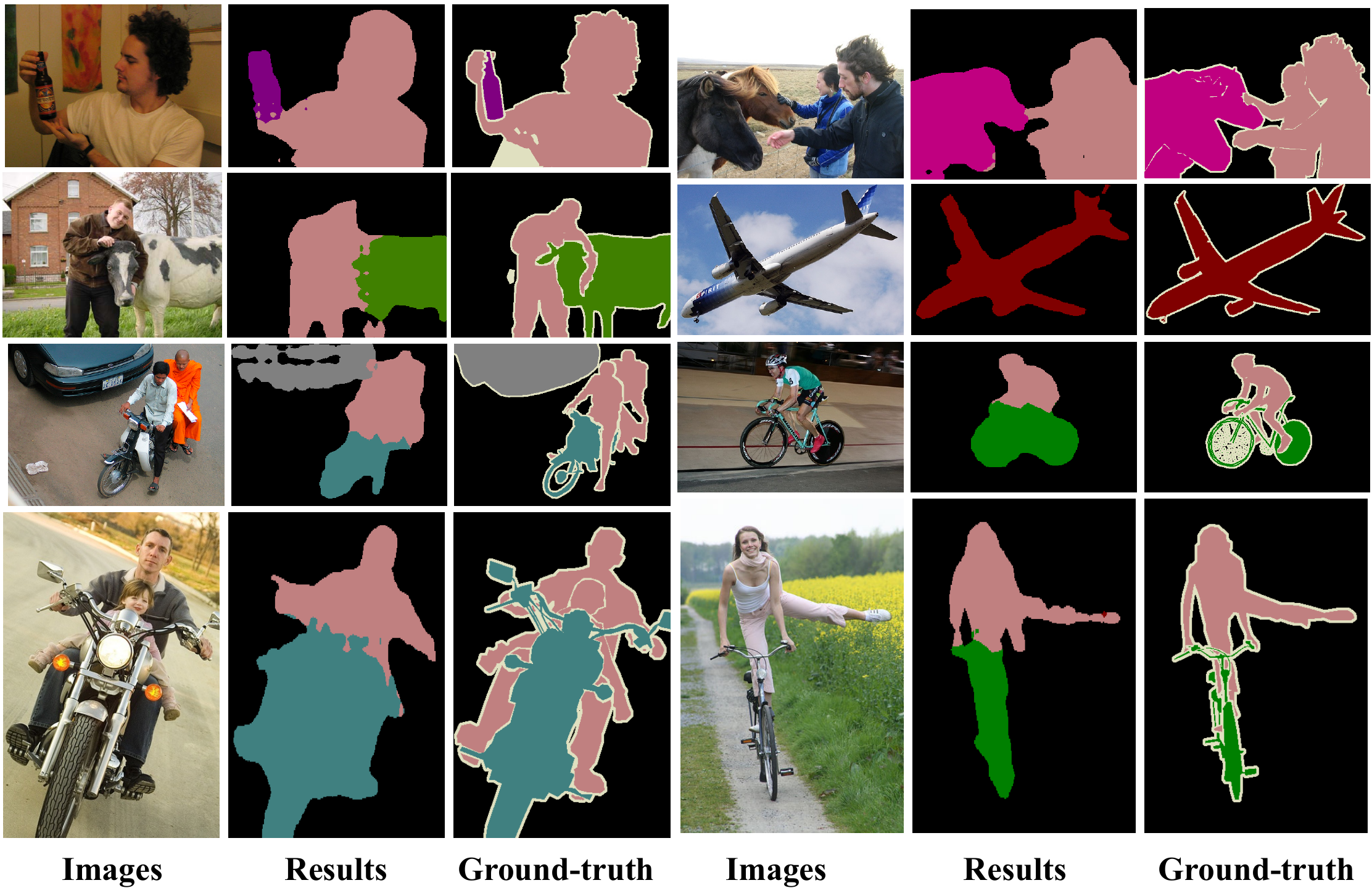}\\
  (a)~~~~~~~~~~~~~~~~~~~~~~~(b)~~~~~~~~~~~~~~~~~~~~~~~(c)
\end{center}
  \caption{Qualitative results of the proposed method on PASCAL VOC 2012 validation set. }
\end{figure}

\subsection{Ablation study}
In this section, we analyze the effect of the proposed jointly learning framework. To validate the impact of multitasking, we show the performance of the networks trained in different single-task and multi-task settings. 

\begin{figure}[htbp]
\begin{center}
  \includegraphics[width=.9\linewidth]{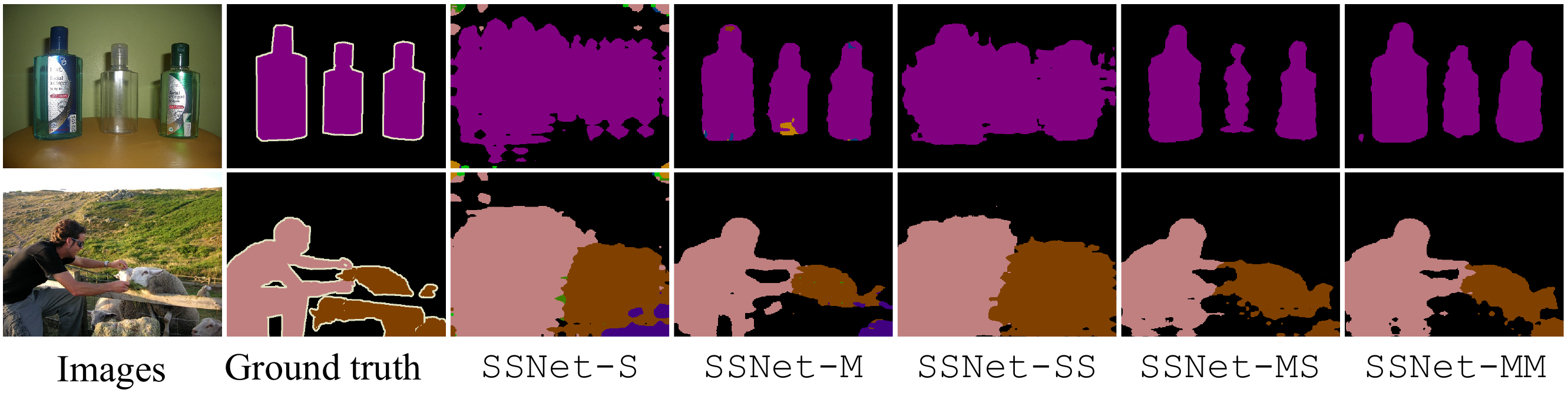}
\end{center}
  \caption{Visual effect of the segmentation results of the models trained in different settings. Images: input images. Ground truth: segmentation ground truth. \texttt{SSNet-S}: results of SSNet-1 trained in single-task setting. \texttt{SSNet-M}: results of SSNet-1 trained in multi-task setting. \texttt{SSNet-S}: results of SSNet-2 trained in single task setting using the predictions of \texttt{SSNet-S}. \texttt{SSNet-MS}: results of SSNet-2 trained in single-task setting usins the predictions of \texttt{SSNet-M}. \texttt{SSNet-MM}: results of SSNet-2 traiend in multi-task setting using the predictions of \texttt{SSNet-M}.}
\label{vis_steps}
\end{figure}

\noindent
\textbf{Semantic segmentation} The quantitative and the qualitative comparison of the models trained in the different settings for segmentation task is shown in Table~\ref{abla_miou} and Figure~\ref{vis_steps} respectively. For the first training stage, we firstly train SSNet-1 in the single-task setting, where only the image-level category labels and $\mathcal{L}_c$ are used. The resulted model is denoted as \texttt{SSNet-S}, of which the mIOU is shown in the first column of Table~\ref{abla_miou}. 
Then we add the saliency task to train SSNet-1 in the multi-task setting. In this setting $\mathcal{L}_c+\mathcal{L}_{s1}$ is used as loss function, with both the image-level category labels and the saliency dataset are used as training data. The resulted model is denoted as \texttt{SSNet-M}, of which mIOU is shown in the second column of Table~\ref{abla_miou}. It can be seen that \texttt{SSNet-M} has a much larger mIOU than \texttt{SSNet-S}, demonstrating that jointly learning saliency detection is of great benefit to WSSS. 
In the second training stage, the training data for SSNet-2 consists of two splits: the one is the predictions of SSNet-1, and the other is the saliency dataset. In order to verify the contribution of each split, we train SSNet-2 in three settings: 1) train with only the predictions of \texttt{SSNet-S} using the $\mathcal{L}_{s2}$ as loss function, 2) train with only the predictions of \texttt{SSNet-M} using the $\mathcal{L}_{s2}$ as loss function, and 3) train with the predictions of \texttt{SSNet-M} and the saliency dataset using $\mathcal{L}_{s1}+\mathcal{L}_{s2}$ as loss function. The resulted models under the three settings is denoted as \texttt{SSNet-SS}, \texttt{SSNet-MS} and \texttt{SSNet-MM}, of which the mIOU scores are shown in the third to the fifth column of Table~\ref{abla_miou}. 
From the comparison of \texttt{SSNet-SS} and \texttt{SSNet-MS}, it can be seen that the model trained with the multi-task setting in the first training stage can provide better training data for the second training stage. The comparison of \texttt{SSNet-MS} and \texttt{SSNet-MM} shows that when trained with the same pixel-level segmentation labels, the model trained in the multi-task setting is still better than the single-task setting. 

\noindent
\textbf{Saliency detection} To study the effect of jointly learning for saliency detection, we compare the performance of SSNet-2 trained in multi-task settings and single-task settings. We firstly train SSNet-2 only for saliency detection task, resulting in a model denoted as \texttt{SSNet-2S}, of which the maximum F-measure and MAE are shown in the first column of Table~\ref{abla_mae}. Then we run the model \texttt{SSNet-MM} mentioned above on saliency dataset, and the resulted F-measure and MAE are shown in the second column of Table~\ref{abla_mae}. As can be seen, the models trained in multi-task setting has a comparable performance to the model trained in single-task setting, where the former has a larger maximum F-measure and the later is better in terms of MAE. Therefore, it is safe to conclude that conduct jointly learning of semantic segmentation dose not harm the performance on saliency detection. This result validates the superiority of the proposed jointly learning framework considering its great benefit to semantic segmentation.
\begin{table}[htbp]
\caption{\small Comparison of the models trained in different settings on semantic segmentation task. \texttt{S} and \texttt{M} represent single-task training and multi-task training respectively. Larger mIOU indicates better performance. The best results are in bold.}
\label{abla_miou}
\small
\begin{center}
\begin{tabular}{c|ccccc}
\hline
training stage & \multicolumn{5}{c}{training strategy}\\
\hline
stage 1 & \texttt S& \texttt M&\texttt S&\texttt M & \texttt M\\
stage 2 & & &\texttt S&\texttt S&\texttt M\\
\hline
mIOU & 33.1 & 57.1 & 47.1 & 62.7 & \textbf{63.3} \\
\hline
\end{tabular}
\end{center}
\vspace{-10pt}
\end{table}
\begin{table}[htbp]
\caption{\small Comparison of the models trained in different settings on saliency detection task (evaluated on ECSSD dataset). \texttt{S} and \texttt{M} represent single-task training and multi-task training respectively. CRF represents the results after CRF post processing. Larger max $F_\beta$ and smaller MAE indicate better performance. The best results are in bold.}
\label{abla_mae}
\small
\begin{center}
{\setlength{\tabcolsep}{1em}
\begin{tabular}{c|ccc}
\hline
  & \multicolumn{3}{c}{training strategy}\\
\hline
&\texttt S &\texttt M&\texttt M\\
\hline
MAE & 0.046 & 0.047& \textbf{0.045} (CRF)\\
max $F_\beta$&0.899 & 0.912 &\textbf{0.919} (CRF)\\
\hline
\end{tabular}}
\end{center}
\vspace{-10pt}
\end{table}

\section{Conclusion}
This paper presents a joint learning framework for saliency detection (SD) and weakly supervised semantic segmentation (WSSS) using a single model, \ie the saliency and segmentation network (SSNet). Compared with WSSS methods exploiting pre-trained SD models, our method makes full use of segmentation cues from saliency annotations and is easier to train. Compared with existing fully supervised SD methods, our method can provide more informative results. Experiments shows that our method achieves state-of-the-art performance among both fully supervised SD methods and WSSS methods. 

\section*{Acknowledgements}
Supported by the National Natural Science Foundation of China  \#61725202, \#61829102, \#61751212, \#61876202, Fundamental Research Funds for the Central Universities under Grant \#DUT19GJ201 and Dalian Science and Technology Innovation Foundation \#2019J12GX039.

{\small
\bibliographystyle{ieee_fullname}
\bibliography{egbib}
}

\end{document}